\documentclass{article}

\usepackage[english]{babel}

\usepackage[a4paper,top=2cm,bottom=2cm,left=3cm,right=3cm,marginparwidth=1.75cm]{geometry}
\usepackage[backend=biber,style=authoryear,citestyle=authoryear-comp,maxcitenames=1,uniquelist=false]{biblatex}
\usepackage{amsmath}
\usepackage{amssymb}
\usepackage{amsfonts}
\usepackage{graphicx}
\usepackage[colorlinks=true, allcolors=blue]{hyperref}
\usepackage{booktabs}
\usepackage{lineno}
\usepackage{wrapfig}
\usepackage{tabularray} % For flexible formatting of tables
\usepackage{microtype} % For formatting em-dash, en-dash, etc.
\usepackage{orcidlink}

\title{SiLVi: Simple Interface for Labeling Video Interactions}
\author{
\parbox{\textwidth}{
\centering Ozan Kanbertay\,\orcidlink{0009-0002-4108-619X}$^{1}$ ,
Richard Vogg\,\orcidlink{0000-0001-9079-2715}$^{1,2}$, Elif Karakoc\,\orcidlink{0009-0008-6196-8847}$^{2}$, Peter M. Kappeler\,\orcidlink{0000-0002-4801-487X}$^{2,3}$, \\
Claudia Fichtel\,\orcidlink{0000-0002-8346-2168}$^{2}$, Alexander S. Ecker\,\orcidlink{0000-0003-2392-5105}$^{1,*}$}}

\begin{document}

\maketitle

{\centering\vspace{-12pt}
\small $^{1}$Institute of Computer Science and Campus Institute Data Science, University of Göttingen \\
\small $^{2}$Behavioral Ecology \& Sociobiology Unit, German Primate Center, Göttingen, Germany \\
\small $^{3}$Department of Sociobiology/Anthropology, University of Göttingen, Göttingen, Germany \\
\small $^*$corresponding author: \texttt{ecker@cs.uni-goettingen.de}
\vspace{12pt}\\
}

%\vspace{-0.9cm}

\begin{abstract}
Computer vision methods are increasingly used for the automated analysis of large volumes of video data collected through camera traps, drones, or direct observations of animals in the wild. While recent advances have focused primarily on detecting individual actions, much less work has addressed the detection and annotation of interactions -- a crucial aspect for understanding social and individualized animal behavior.
Existing open-source annotation tools support either behavioral labeling without localization of individuals, or localization without the capacity to capture interactions. To bridge this gap, we present SiLVi, an open-source labeling software that integrates both functionalities. SiLVi enables researchers to annotate behaviors and interactions directly within video data, generating structured outputs suitable for training and validating computer vision models. By linking behavioral ecology with computer vision, SiLVi facilitates the development of automated approaches for fine-grained behavioral analyses. Although developed primarily in the context of animal behavior, SiLVi could be useful more broadly to annotate human interactions in other videos that require extracting dynamic scene graphs.
The software, along with documentation and download instructions, is available at: \href{https://silvi.ozankanbertay.com}{https://silvi.ozankanbertay.com}.
   
\end{abstract}

\begin{figure}[!ht]
    \centering
    \vspace{12pt}
    \makebox[\textwidth]{%
        \includegraphics[width=1.2\linewidth]{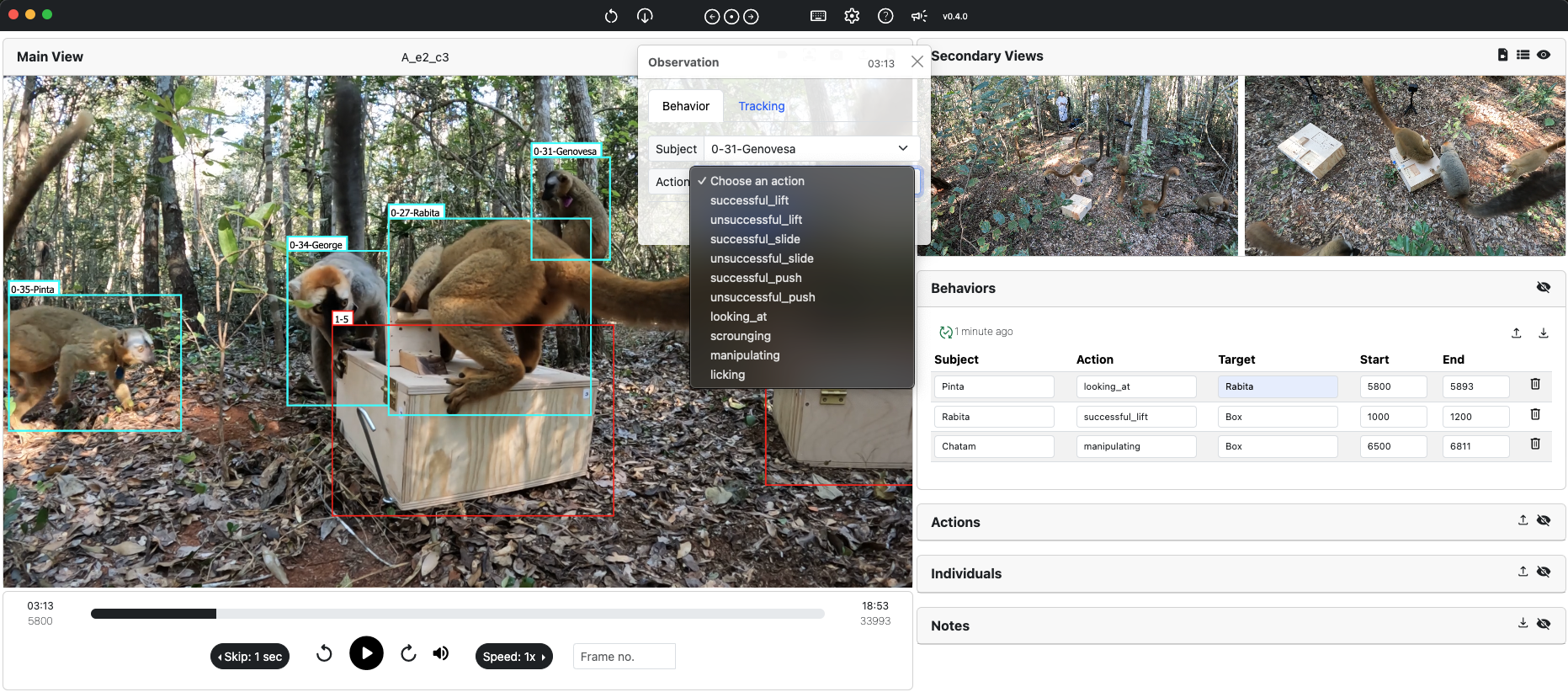}
    }
    \caption{Scoring the behavior of redfronted lemurs using SiLVi: Users can upload multiple video views, an ethogram, tracking and individual identification files. The app allows for annotation of actions and interactions, as well as fast correction of tracking and identification errors. }
    \label{fig:enter-label}
\end{figure}

\section{Introduction}

Video recordings are an important tool in behavioral ecology, providing a non-invasive and scalable means to observe animal behavior over extended periods and across diverse environments that is more comprehensive than focal sampling. Behavioral ecologists use video footage to study individual actions and social interactions in naturalistic settings, but also to document experiments. Manual annotation of such videos has long been a cornerstone of behavioral analyses, allowing researchers to quantify complex behaviors that are difficult to capture in real time. However, the manual nature of this process is labor-intensive, time-consuming, and subject to observer bias.

The growing availability of annotated, open-source datasets such as PanAf20k \parencite{brookes2024panaf20k}, ChimpAct \parencite{ma2023chimpact}, ChimpBehave \parencite{fuchs2025forest}, and BaboonLand \parencite{duporge2025baboonland} has enabled significant progress in the development of machine learning models for the automated detection of temporally and spatially localized actions in video data \parencite{ma2024alphachimp, brookes2024chimpvlm, goss2024quantifying}. These advancements have the potential to enable researchers to automate parts of the annotation process, offering the possibility of greater consistency and scalability.

Nonetheless, many important behavioral phenomena, particularly interactions between individuals, remain difficult to model. Interactions are a critical component of social behavior and are often central to understanding dominance structures, cooperation, communication, and other complex social dynamics. Current machine learning approaches for primate behavior typically decompose dyadic or group interactions into simultaneous individual actions, for example ``grooming'' and ``being groomed'' \parencite{duporge2025baboonland, ma2023chimpact}. While this may work for simple or physically close interactions, it introduces limitations. For instance, when multiple grooming events occur simultaneously within a group, the pairwise associations become ambiguous. Moreover, interactions involving spatial distance, such as one individual looking at another, cannot be readily captured through decomposed individual actions.

One key obstacle to progress in this area is the lack of annotated datasets that explicitly capture interactions with spatial and temporal precision. 
Although existing annotation tools such as BORIS \parencite{friard2016boris} support the labeling of social behaviors, they typically do so in a non-localized manner, marking only the presence of an interaction over time rather than its spatial extent within the video frame. This limitation prevents their use in training deep learning models that rely on spatially localized input.
Some tools allow for spatio-temporal annotations, but these are restricted to actions rather than interactions \parencite{cvat, dutta2019via, price2025framework, elhorst2025behave}. Other tools do support spatio-temporal annotation of interactions \parencite{chen2023alphatracker, segalin2021mouse}; yet, as they are based on keypoints, they are better suited to applications with static laboratory backgrounds, since keypoint tracking remains challenging in more complex settings \parencite{perez2023cnn, vogg2024primat}.

To address this gap, we introduce SiLVi, a \textit{\textbf{S}imple \textbf{I}nterface for \textbf{L}abelling \textbf{V}ideo \textbf{I}nteractions}. SiLVi is a lightweight, open-source tool designed specifically to facilitate the annotation of interactions in a way that is both spatially and temporally localized, enabling the generation of training data for computer vision models. By supporting precise annotations of both actions and interactions across individuals, SiLVi empowers researchers to move beyond individual-centric behavioral models toward richer representations of social behavior. Apart from this main contribution, SiLVi also allows for labeling individual IDs and annotating or correcting tracks. 

\section{SiLVi: Simple Interface for Labeling Video Interactions}
\label{sec:silvi}

SiLVi is a user-friendly software that enables users to annotate actions and interactions on videos.
It is currently available for macOS, Windows, and Linux operating systems. A web version is also planned to be released soon. The app is written with JavaScript using the Electron framework to make it available on multiple platforms, including the web, with a single codebase. Updates are provided automatically.

SiLVi supports common video formats --- including MP4 and WebM --- and codecs. Codecs H.264 and VP8/VP9 are universally supported, while H.265 and AV1 support depends on device hardware and operating system. This ensures broad compatibility with video recordings from smartphones, cameras, and browser-based tools.

\paragraph{Navigation.}   
If an experiment includes multiple video recordings of the same scene with different points of view, the most important video should be loaded into ``Main View'', while additional views can be displayed side by side on a single application window (Fig.~\ref{fig:enter-label}, ``Secondary Views'' at the top right). When a secondary view is imported into the application, its playback is automatically synchronized with the main view. Capabilities for moving faster or slower, skipping a given number of seconds backwards or forward, or moving frame-by-frame exist (Fig.~\ref{fig:enter-label}, buttons below ``Main View''). 
Manual input of the desired target frame number is also possible. The time bar shows mini thumbnails when hovering over it.
The main view enables users to zoom in on regions of interest, which can be helpful to detect precise boundaries of an object or detailed interactions.
During the whole annotation process, users can write free-form notes that will be saved with the video (Fig.~\ref{fig:enter-label}, bottom right corner).

\paragraph{User input.}
To create spatially localized annotations of interactions, a tracking file for the main video can be imported into the app. The tracking file should either be in simplified or extended MOT Challenge format \parencite{dendorfer2020motchallenge}. The simplified format consists of a text file with one line per detection containing frame number, object ID, bounding box coordinates (with the top left corner of the bounding box, width and height), detection confidence and object class. Alternatively, the extended format additionally includes individual ID and confidence. Such a tracking file can be obtained by processing a video with a multi-animal tracking model such as PriMAT \parencite{vogg2024primat}. Animal detection and tracking are not performed by SiLVi but are assumed to be a separate pre-processing step. Tracking files are displayed as bounding boxes, each with an object class and ID. These bounding boxes can be hidden if they obstruct the view. Additionally, the opacity, thickness, and color of bounding boxes for each class can be adjusted to enhance the viewing experience. Apart from the track ID, the individual ID will be displayed if the tracking file is in extended MOT Challenge format. To display individuals' names, users can upload a single CSV file with one individual name per line into the panel ``Individuals'' (Fig.~\ref{fig:enter-label}, right). This will automatically translate the individual ID from the tracking file into the respective name. Similarly, the ethogram can be uploaded with a CSV file into the panel ``Ethogram''. If the dataset contains fewer than 26 individuals or actions, the app automatically assigns keyboard shortcuts, one for each letter of the alphabet, prioritizing letters that match the initial letter of each name when possible. These automatically generated shortcuts can be changed at any time in the settings. In case of more than 26 individuals or actions, the app will assign shortcuts only to the first 26 of them. The rest of the shortcuts can be determined by the user utilizing key combinations.

\paragraph{Correcting tracking and identification labels.}
The tracking and identification file can contain errors such as imprecise detections or ID switches during tracking. SiLVi allows for fast and easy corrections of different types of error (see Table~\ref{tab:errors}). 
Users can choose to apply corrections to an entire track, to a track after a given frame, or only to the current frame. Correcting tracking errors during annotation increases precision in observations and enables researchers to improve tracking models by training them with higher-quality data. Tracks can optionally be annotated with individual IDs by right-clicking on a bounding box and selecting the individual's name from a drop-down menu. The name will then be propagated throughout the whole track.

\paragraph{Interaction annotation.}
SiLVi makes the bounding boxes in the tracking file interactive for annotating interactions and actions (Fig.~\ref{fig:enter-label}, ``Main View''). 
Annotations can be done by either clicking on the bounding boxes or utilizing the customizable keyboard shortcuts assigned to subjects and items in the ethogram. The difference between interactions and actions is that interactions are annotated with a subject and a target, while actions are directly tagged to a subject. The ``Behaviors'' panel (Fig.~\ref{fig:enter-label}, right) lists all annotated interactions and actions with respective start and end times. By clicking on the start or end time of an interaction, the video jumps to the precise frame. In case of errors during annotation, all fields can be manually corrected. To prevent typos, the app restricts individuals and actions to options from their respective predefined lists and provides auto-completion by default. 

\paragraph{Data export.}
Annotations for tracking, identification, action detection, as well as free-form notes for the video, can be exported in CSV format. Tracking and identification data are exported in the same format as they were input. Identification files contain subject, (inter)action, target, start, end and duration of the interaction both in frames and in seconds. Additionally, metadata can be exported, including the annotator’s name (required when opening the app for the first time and can be changed later in the “Settings” menu), the video name, and the available individuals and interactions. All exported files automatically get the name of the video in the main view, and a suffix describing what annotations the file contains (e.g. ``tracking'', ``metadata''). 
SiLVi saves the current status of the annotation at all times and resumes at the current state of the annotation, even after closing and restarting. Users can take snapshots of images within the videos, which are stored with the video name and the extracted frame number. The snapshots can be saved with or without bounding box annotations.

\begin{table}[b!]
\centering
\caption{Common tracking and identification errors and how to fix them.}
\begin{tblr}{
    colspec = { X[0.9
    ,l,m] |X[1.4,l,m]| X[2.4,l,m] },
    hlines, vlines,
}
\textbf{Type of Error} & \textbf{Example} & \textbf{How to Fix} \\
Misclassification & A lemur is labeled with class ``feeding box''. & Click on the track and go to the tracking panel. Select ``lemur'' from the drop-down menu in the ``Change class'' option. \\
Misidentification & The individual named George is assigned the ID Genovesa. & Right-click on the track and select George from the drop-down menu. \\
ID switch & The track starts on the correct individual, but continues on a different individual. & Click on the track on the frame where the ID switch is happening and go to the tracking panel. Select ``Assign the next unused ID'' and choose ``All frames after this frame''. \\
False positive detection & A bounding box is produced at a location without an object. & Click on the track, go to the tracking panel. Select ``Remove track''. \\
False negative detection & An object is not detected. & Enable drawing mode and draw a bounding box around the undetected individual. If the movement trajectory is linear, go to the last frame of the movement, draw a new bounding box, and select ``Interpolate''. \\
Partial detection & An object is detected but the bounding box is not precise. & Enable drawing mode and resize the bounding box by dragging the handles on its corners. \\
\end{tblr}
\label{tab:errors}
\end{table}

\section{Example of use}

\begin{wrapfigure}[20]{r}{0.5\textwidth}
    \centering
    \vspace{-12pt}
    \includegraphics[width=\linewidth]{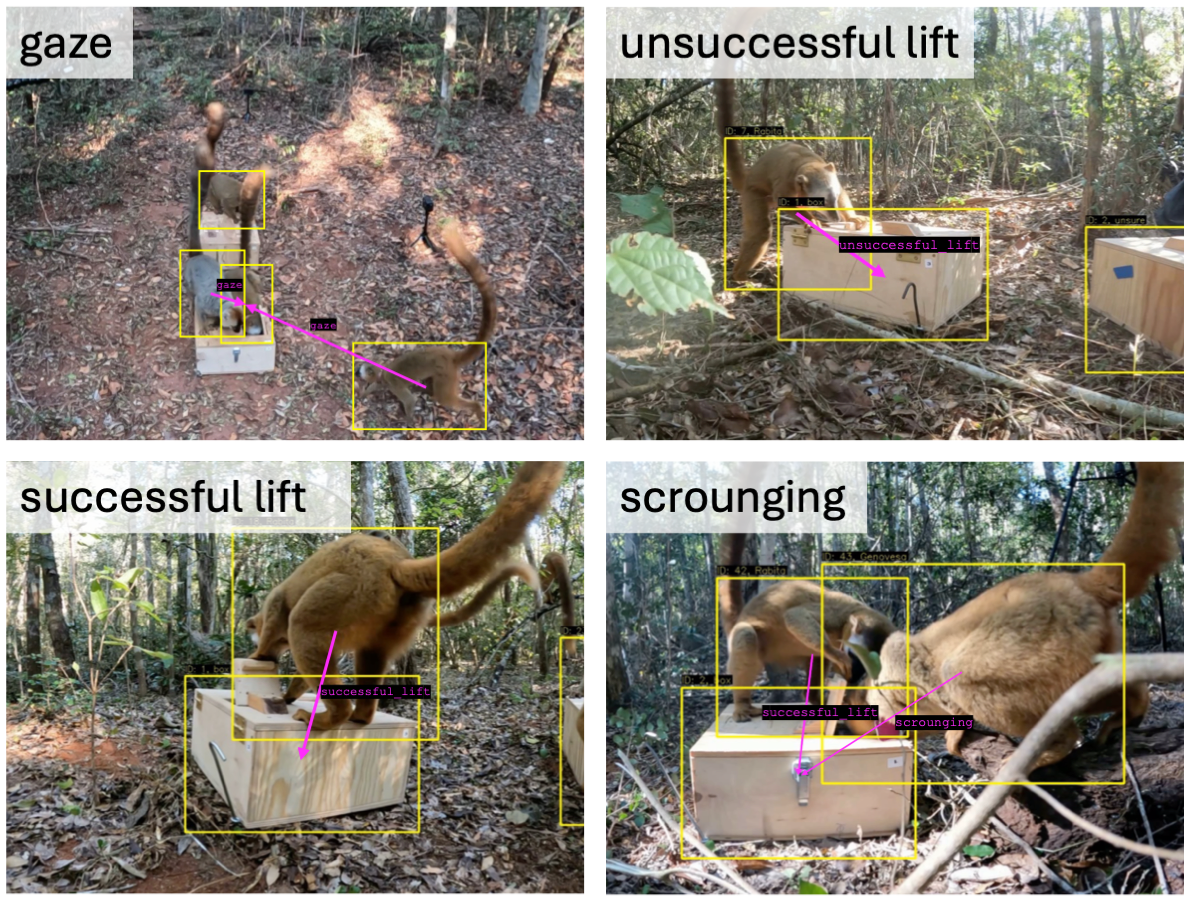}
    \caption{Examples of different types of interaction. Gaze can be detected on single images, while the interactions with the feeding box often require temporal context.}
    \label{fig:examples}
\end{wrapfigure}

We tested the app with videos of redfronted lemurs (\textit{Eulemur rufifrons}) in the wild. The setup of the experiments with eight cameras filming the lemurs during social learning experiments in Kirindy Forest, Madagascar, described in detail by \textcite{karakoc2025foraging}. In these experiments, red-fronted lemurs were able to learn how to open food boxes using two different techniques: lifting the lid or pushing the box backwards to get to the food inside. While \textcite{karakoc2025foraging} used BORIS \parencite{friard2016boris} for manual annotations on 80 hours of video material, their recent data collection involved even larger data volumes (far over 650 hours), and made the use of computer vision models necessary. 
The multiple views were temporally synchronized using an auditory signal and loaded into the app. The ethogram contained ten interaction classes: \emph{looking at}, \emph{successful lift}, \emph{unsuccessful lift}, \emph{successful push}, \emph{unsuccessful push}, \emph{successful slide}, \emph{unsuccessful slide}, \emph{scrounging}, \emph{licking} and \emph{manipulating the food box} (see examples in Fig.~\ref{fig:examples}). 
Interactions with food boxes could alternatively be labeled as actions by simply tagging the activity to the bounding box of the individual performing the action. In this case, users can skip the target, and only annotate subject and action. In total, four groups of lemurs were studied in group sizes from seven to twelve individuals, most of which participated in the experiments. For each group, we created one file with the names of all individuals, also including the option \emph{Unsure} in case individuals were not identifiable. 

While gaze was annotated on videos filmed with a wider angle, the interactions with the feeding boxes were labeled on close-up views. In addition to the main view, we always opened two views for complementary information. 

Annotators used keyboard shortcuts for labeling behaviors, as these were faster than clicking on bounding boxes. Keyboard shortcuts also allow annotators to continue labeling even when the tracking model temporarily fails to detect an individual, because actions are assigned directly to an individual’s name rather than to a bounding box. As a result, post-processing was required to check on which frames tracking, identification, and behavior annotations are aligned. Any video segments in which the tracking failed or identities could not be reliably maintained were filtered out from the dataset used for model training. 

In total, we annotated feeding box interactions on 176 videos and gaze on 50 videos, amounting to 56 hours of video footage.

\section{Discussion and future directions}
This \textit{Simple Interface for Labeling Video Interactions (SiLVi)} allows researchers to annotate spatio-temporal interactions and actions.
As computer vision methods are increasingly used in automated video annotation, for behavioral ecology \parencite{vogg2025computer}, in wildlife conservation or monitoring of animal movement \parencite{kholiavchenkoDeepDiveKABR2024}, we need tools that facilitate the annotation of behavior in a suitable format for model development.

Behavioral ecologists can use SiLVi to manually annotate behaviors in their video recordings. Computer vision researchers can utilize the annotated bounding boxes, identification labels, actions, and interactions to train models for automated detection of behavior in both laboratory settings and the wild. \parencite{vogg2024primat, liu2021fully, jiangVrdONEOnestageVideo2024, ryan2025gaze} \parencite{vogg2024primat, liu2021fully, jiangVrdONEOnestageVideo2024, ryan2025gaze}. This enables automated processing of large volumes of video data captured by camera traps, drones, or manual recording.

The interface is intentionally designed to be simple and intuitive, ensuring accessibility for annotators with diverse backgrounds. However, this design limits customization beyond the predefined use cases. Nevertheless, the open-source nature of the codebase enables developers to extend and adapt the application to meet specific requirements.

Currently, SiLVi is designed for manual interaction annotation and does not include built-in computer vision assistance. Features such as automated bounding box proposals or individual identification must be generated externally using dedicated tools --- e.g., PriMAT \parencite{vogg2024primat} --- and subsequently imported into SiLVi, as described in Section \ref{sec:silvi}. To improve this workflow, we are exploring the integration of tracking and identification models into , which users will be able to optionally enable.

SiLVi is a versatile tool for manual behavioral annotation, applicable to a wide range of video data beyond animal behavior research. Its flexible design supports diverse domains requiring fine-grained annotation, including human–computer interaction, clinical and sports performance analysis, and social or educational research. By enabling precise and reproducible annotation across diverse contexts, SiLVi provides a valuable resource for researchers studying complex behavioral dynamics in naturalistic or controlled environments.

\section*{Acknowledgments}

This project was funded by the Deutsche Forschungsgemeinschaft (DFG, German Research Foundation) via project number 454648639 – SFB 1528.
We thank student assistants Joana Niedner, Lucia Fricke, Lea Hoffmann, and Solene Joubert for their assistance in annotating the video data and providing valuable feedback on the application.

\section*{Author contributions}

O.K. developed the software and wrote the documentation. O.K. and R.V. led the manuscript writing. R.V. and A.E. conceived the project and provided overall supervision. R.V. designed the initial software prototype. E.K., C.F., and P.K. collected the data, supervised the annotation process for the use case, and provided feedback during software development. A.E., C.F., and P.K. secured funding. All authors critically reviewed and contributed to the manuscript.

\printbibliography
%\bibliographystyle{apalike}
%\bibliography{refs}

\end{document}